\documentclass[11pt,a4paper]{article}

\usepackage[utf8]{inputenc}
\usepackage[T1]{fontenc}
\usepackage{amsmath,amssymb,amsfonts}
\usepackage{graphicx}
\usepackage{booktabs}
\usepackage{hyperref}
\usepackage{xcolor}
\usepackage{algorithm}
\usepackage{algorithmic}
\usepackage{multirow}
\usepackage{caption}
\usepackage{subcaption}
\usepackage{geometry}
\usepackage{natbib}
\usepackage{listings}
\usepackage{tikz}

\title{%
  \textbf{Playing DOOM with 1.3M Parameters:\\Specialized Small Models vs Large Language Models\\for Real-Time Game Control}
}

\author{
  David Golchinfar$^{1}$ \and Daryoush Vaziri$^{2}$ \and Alexander Marquardt$^{3}$ \\\\
  $^{1}$VAGO Solutions, Germany \\
  $^{2}$University of Applied Sciences Bonn-Rhein-Sieg, Germany \\
  $^{3}$Cybernetics and Reality Engineering (CARE) Laboratory, \\
  Nara Institute of Science and Technology, Ikoma, Japan
}

\date{\today}

\begin{document}

\maketitle

\begin{abstract}
We present SauerkrautLM-Doom-MultiVec, a 1.3 million parameter model that plays the classic first-person shooter DOOM in real time, outperforming large language models up to 92,000$\times$ its size, including Nemotron-120B, Qwen3.5-27B, and GPT-4o-mini. Our model combines a ModernBERT encoder with hash embeddings, depth-aware token representations, and an attention pooling classification head to select game actions from ASCII frame representations at 31ms per decision. Trained on just 31,000 human gameplay demonstrations, it achieves 178 frags in 10 episodes (17.8 per episode) in the \texttt{defend\_the\_center} scenario, more than all tested LLMs combined (13 frags total). All agents receive equivalent input: ASCII frames and depth maps. Despite having 92,000$\times$ fewer parameters than Nemotron-120B, our model is the only agent that actively engages enemies rather than purely evading them. These results demonstrate that small, task-specific models trained on domain-appropriate data can decisively outperform general-purpose LLMs at real-time control tasks, at a fraction of the inference cost, with deployment capability on consumer hardware.
\end{abstract}

\section{Introduction}
\label{sec:intro}

The rapid scaling of large language models (LLMs) has led to remarkable generalization across tasks, from code generation to mathematical reasoning. Models like GPT-4o, Claude, and Gemini demonstrate capabilities that were unimaginable a few years ago. However, this generalization comes at enormous cost: billions of parameters, specialized datacenter hardware for inference, API latency measured in seconds, and per-query pricing that makes real-time applications prohibitive.

We argue that for many practical tasks, particularly those requiring real-time decision-making, edge deployment, or domain-specific reasoning, small specialized models remain not just competitive but \textit{superior} to general-purpose LLMs. To demonstrate this, we present an extreme case study: teaching a 1.3 million parameter model to play DOOM, and benchmarking it against state-of-the-art multimodal LLMs, all of which support vision input and could, in principle, process raw game screenshots natively.

Our contributions are:
\begin{enumerate}
    \item \textbf{ModernBERT-Hash}: A parameter-efficient encoder combining ModernBERT's architecture with hash embeddings, reducing embedding parameters by 73\% while maintaining representation quality.
    \item \textbf{Depth-aware ASCII encoding}: A novel game state representation that combines ASCII art with quantized depth embeddings, enabling a text-based model to perceive 3D spatial structure.
    \item \textbf{Human demonstration training}: An efficient data collection pipeline using VizDoom's spectator mode, producing high-quality training data with depth annotations.
    \item \textbf{Comprehensive LLM benchmark}: A fair comparison against Nemotron-120B (120B params), Qwen3.5-27B (27B params), GPT-4o-mini, and Gemini Flash Lite, all receiving equivalent ASCII + depth input.
\end{enumerate}

Our 1.3M parameter model achieves 178 frags in 10 episodes (17.8 per episode) while all tested LLMs achieve only 13 combined, despite some having nearly 100,000 times more parameters.

\section{Related Work}
\label{sec:related}

\subsection{Small Models vs.\ Large Language Models}

The efficiency advantages of task-specific models over general LLMs have been documented across domains. \citet{schick2021fewshot} demonstrated that small models with task-specific training can match GPT-3's performance at a fraction of the cost. Microsoft's Phi-3 \citep{abdin2024phi3} showed that a 3.8B model rivals GPT-3.5 through data quality alone. \citet{taskspecific2026} introduced the Performance-Efficiency Ratio metric, finding that small models (0.5--3B) achieve superior efficiency across all tested natural language processing (NLP) tasks. \citet{belcak2025slm} argue that small language models are the future of agentic AI, delivering 10--30$\times$ cost savings. The TinyML movement \citep{banbury2021mlperf} has shown that models under 1MB can run meaningful inference on microcontrollers. Our work pushes this to the extreme: a sub-2M parameter model outperforming trillion-parameter LLMs on a real-time control task.

\subsection{ModernBERT and Efficient Transformers}

ModernBERT \citep{modernbert2024} modernizes the BERT architecture with rotary position embeddings (RoPE), alternating local and global attention layers, and native Flash Attention 2 support \citep{dao2023flashattention2}. The local attention pattern with sliding windows of 128 tokens is particularly relevant for our application, where spatial locality in ASCII frames maps directly to spatial locality in the game world. Hash embeddings \citep{svenstrup2017hash}, building on the feature hashing principle \citep{weinberger2009feature} and neural network compression via hashing \citep{chen2015hashing}, reduce the embedding table from $V \times H$ to $V \times P + P \times H$ parameters, where $P \ll H$ is the projection dimension. NeuML's BERT-Hash models \citep{neuml2025berthash} demonstrated this approach on the original BERT architecture, producing sub-1M parameter models (e.g., bert-hash-nano with 969K parameters) that achieve 99\% embedding parameter reduction. We advance this line of work to ModernBERT \citep{modernbert2024}, which brings rotary position embeddings, alternating local/global attention, and native Flash Attention 2 support, resulting in a more capable encoder within the same parameter budget.

\subsection{Game AI and Neural Game Engines}

VizDoom \citep{wydmuch2019vizdoom} provides a research platform for visual reinforcement learning (RL) in first-person shooter environments. \citet{lample2017playing} trained agents using deep RL with game features, while \citet{dosovitskiy2017learning} used direct visual input. Most recently, GameNGen \citep{valevski2024gamengen} demonstrated a neural network that can generate DOOM gameplay frames in real time, using diffusion models trained on 70M Proximal Policy Optimization (PPO) agent frames. Our approach differs fundamentally: rather than generating frames, we classify actions from ASCII representations, requiring six orders of magnitude fewer parameters.

\subsection{LLMs for Game Playing}

Several works have explored using LLMs as game agents. \citet{wang2023voyager} used GPT-4 as a Minecraft agent through code generation, while \citet{tsai2023textgames} found LLMs struggle with spatial reasoning even in purely text-based games. \citet{waytowich2024atarigpt} benchmarked multimodal LLMs (GPT-4V, Gemini) on Atari games and found them incapable of zero-shot gameplay. \citet{gallotta2024llmgames} survey the landscape and identify real-time spatial games as fundamentally unsuited for LLM control. Critically, \citet{asciibench2024} showed that standard LLM tokenizers destroy the 2D spatial coherence of ASCII art. Our character-level tokenizer explicitly addresses this by preserving one-to-one character-to-token mapping.

\subsection{Late Interaction and Multi-Vector Models}

ColBERT \citep{khattab2020colbert} introduced late interaction for information retrieval, where per-token embeddings enable fine-grained matching via MaxSim scoring. PyLate \citep{pylate2025} provides a training framework for such models. While we initially attempted ColBERT-style MaxSim for action classification, we found that the sparse gradient signal from the max operator causes embedding collapse in classification settings (Section~\ref{sec:evolution}). Our final architecture preserves the multi-vector encoder but uses attention pooling for classification.

\section{Method}
\label{sec:method}

\subsection{Problem Formulation}

Given a DOOM game frame $F \in \mathbb{R}^{H \times W \times 3}$ and its associated depth buffer $D \in \mathbb{R}^{H \times W}$, we seek a function $f_\theta$ that maps to an action distribution over $K=4$ discrete actions: \texttt{shoot}, \texttt{move\_forward}, \texttt{turn\_left}, \texttt{turn\_right}. The function must execute in under 35ms to maintain real-time gameplay at DOOM's native 35 frames per second (FPS).

\subsection{ASCII + Depth Encoding}

We convert each game frame to a dual representation. The choice of ASCII as the primary encoding is motivated by three properties: (1) it is natively compatible with text encoders without requiring a vision backbone, (2) the 2D character grid preserves spatial layout (row 0 is the top of the screen, column 0 is the left edge), enabling spatial reasoning through character positions, and (3) it achieves extreme compression: a $640 \times 480 \times 3$ RGB frame (921 KB) becomes a 1,024-character string (${\sim}$1 KB), a 900$\times$ reduction with sufficient information for action selection.

\paragraph{ASCII Frame.} The RGB frame is converted to grayscale, downscaled to $40 \times 25$ via block averaging, and each pixel mapped to one of 10 brightness characters: \texttt{" .:-=+*\#\%@"}, ordered from dark (distant/empty space) to bright (nearby solid objects). Row separators (\texttt{\textbackslash n}) produce a 1,024-character string. Bright characters like \texttt{\#} and \texttt{@} typically indicate nearby walls or enemies; dark characters like \texttt{.} and \texttt{:} indicate distant or empty areas. This brightness gradient provides a coarse depth signal even without explicit depth data.

\paragraph{Depth Embedding.} While ASCII brightness correlates with distance, it is insufficient for reliable depth perception: a bright character could be a nearby wall or a distant bright light source. We address this with explicit depth embeddings. The VizDoom depth buffer is downscaled to match the ASCII resolution ($40 \times 25$), normalized to $[0, 1]$, and quantized into 16 discrete bins (i.e., equal-width intervals that partition the continuous depth range into 16 distance categories from near to far). An additional 17th embedding is reserved for special tokens (\texttt{[CLS]}, \texttt{[PAD]}) that have no associated depth value, yielding 17 depth embeddings in total (Table~\ref{tab:params}). Each bin has a learned embedding vector of dimension $H=128$, which is \textit{added} to the corresponding token embedding, analogous to positional embeddings but encoding distance rather than position:

\begin{equation}
    \mathbf{e}_i = \text{HashEmb}(\text{token}_i) + \text{DepthEmb}(\text{depth\_bin}_i)
\end{equation}

This fusion means each token carries both \textit{what} it represents (via the character embedding) and \textit{how far away} it is (via the depth embedding). The 16-bin quantization provides approximately 40cm resolution at typical DOOM engagement distances, which is sufficient to distinguish ``enemy at melee range'' from ``enemy across the room.''

\begin{figure*}[t]
\centering
\includegraphics[width=\textwidth]{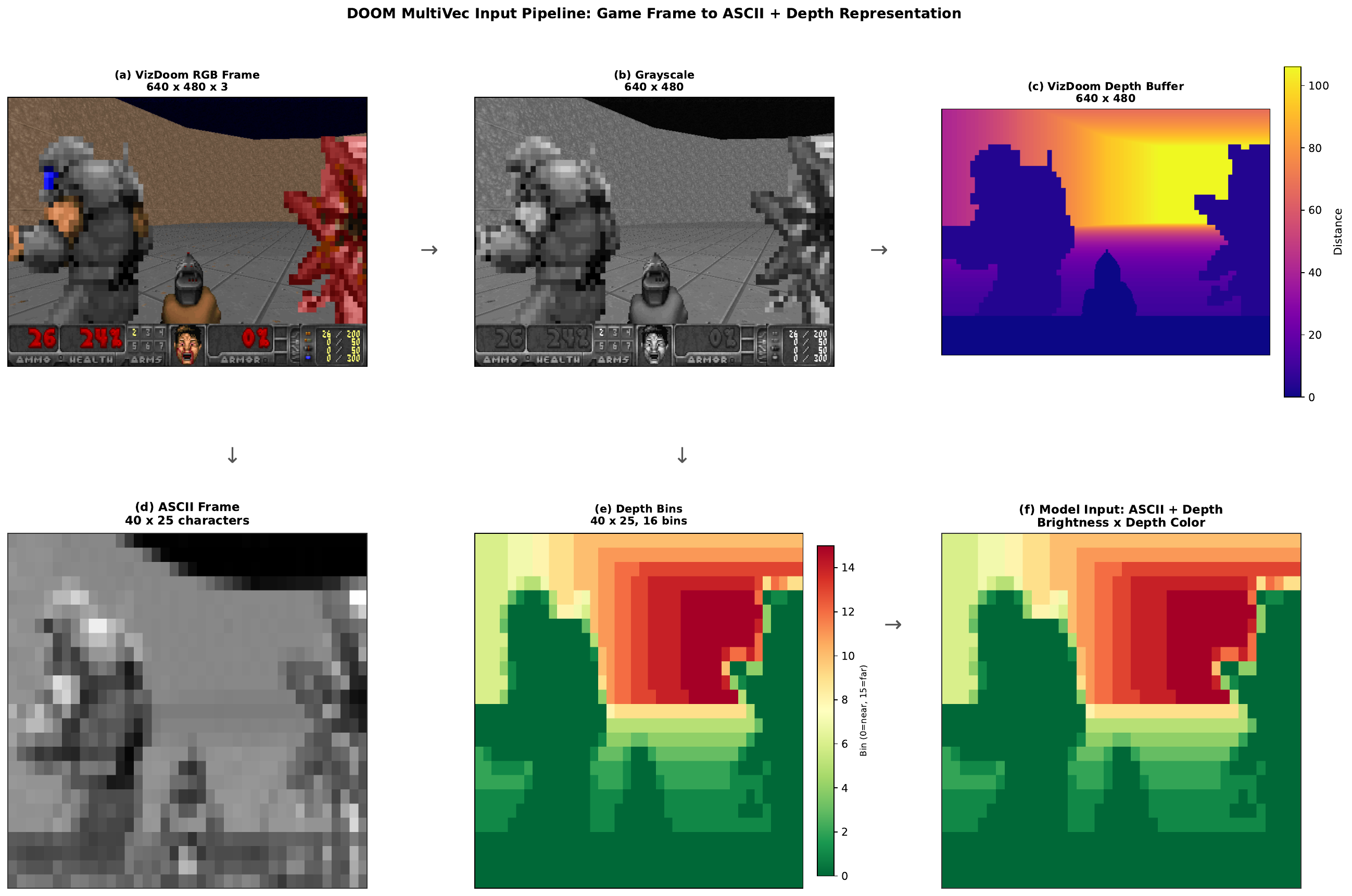}
\caption{Input pipeline: a VizDoom game frame is transformed into the model's dual input representation. (a)~The original 640$\times$480 RGB frame showing enemies at varying distances. (b)~Grayscale conversion. (c)~VizDoom's per-pixel depth buffer (brighter = farther). (d)~The 40$\times$25 ASCII frame rendered as a brightness map, where each pixel corresponds to one of 10 brightness characters. (e)~Depth bins: the depth buffer quantized to 16 bins at ASCII resolution, showing enemy silhouettes (red = near, green = far). (f)~The combined model input: ASCII brightness modulated by depth color. The model receives both representations as token-level embeddings that are summed before encoding.}
\label{fig:pipeline}
\end{figure*}

\subsection{Model Architecture}

\subsubsection{ModernBERT-Hash Encoder}

Our encoder is a 5-layer ModernBERT model with the following modifications:

\paragraph{Hash Embeddings.} Instead of a standard embedding table ($V \times H = 128 \times 128 = 16{,}384$ parameters), we use a two-stage hash embedding \citep{svenstrup2017hash}: a compact lookup ($V \times P = 128 \times 16 = 2{,}048$) followed by a linear projection ($P \times H = 16 \times 128 = 2{,}048$) and LayerNorm. This approach draws on the feature hashing framework \citep{weinberger2009feature, chen2015hashing} where multiple inputs share compressed representations through hash-based indexing. Total embedding parameters: 4,480 (vs.\ 16,384 standard), a 73\% reduction. With our tiny 75-token ASCII vocabulary (69 character tokens + 6 action tokens), this savings is modest in absolute terms but demonstrates the architecture's scalability to larger vocabularies.

\paragraph{Local + Global Attention.} Following ModernBERT, we alternate between local attention (sliding window of 128 tokens) and global attention (every 3rd layer). With 5 layers, layers 0 and 3 use global attention while layers 1, 2, and 4 use local attention. This is well-suited to ASCII frames where local spatial patterns (adjacent characters forming walls, enemies, or corridors) are as important as global context (where am I relative to the arena center?).

\paragraph{Character-Level Tokenizer.} Standard Byte Pair Encoding (BPE) tokenizers used by LLMs are destructive for ASCII art: they merge common character sequences into single tokens, collapsing spatial structure. As \citet{asciibench2024} demonstrated, this destroys the 2D coherence that makes ASCII art interpretable. For example, a BPE tokenizer might merge the sequence \texttt{...} into a single token, losing the information that these are three distinct spatial positions in the frame.

We use a character-level tokenizer with 75 tokens: 10 brightness characters, 7 entity markers, 26 lowercase letters, 10 digits, 10 punctuation marks, 6 action-specific special tokens (\texttt{[ACT\_SHOOT]}, etc.), and 5 standard special tokens (\texttt{[PAD]}, \texttt{[CLS]}, etc.). No subword merging is performed. Each ASCII character maps to exactly one token, preserving a strict one-to-one mapping between spatial positions in the game frame and token positions in the sequence. A $40 \times 25$ frame with newline separators produces exactly 1,024 content tokens, each corresponding to a unique spatial coordinate.

\subsubsection{Attention Pooling Classification Head}

The encoder produces per-token embeddings $\mathbf{H} \in \mathbb{R}^{S \times 128}$ where $S \approx 1{,}024$. We aggregate these via learned attention pooling:

\begin{equation}
    \alpha_i = \frac{\exp(\mathbf{w}^\top \mathbf{h}_i)}{\sum_j \exp(\mathbf{w}^\top \mathbf{h}_j)}, \quad
    \mathbf{v} = \sum_i \alpha_i \mathbf{h}_i
\end{equation}

where $\mathbf{w} \in \mathbb{R}^{128}$ is a learned attention vector. The pooled representation $\mathbf{v}$ is classified by a linear layer: $\mathbf{y} = \mathbf{W}_c \mathbf{v} + \mathbf{b}_c$ where $\mathbf{W}_c \in \mathbb{R}^{4 \times 128}$.

We chose attention pooling over mean pooling and CLS token aggregation based on empirical evaluation. Mean pooling treats all tokens equally, diluting the signal from game-relevant tokens (enemies, walls) with background tokens (empty space). CLS pooling relies on a single special token to attend to all positions, which is difficult to learn from 31K samples. Attention pooling provides a middle ground: the learned attention vector $\mathbf{w}$ assigns high weights to tokens in informative regions (e.g., enemy characters near the center of the frame) and low weights to background, enabling the model to focus on the spatial features most relevant for action selection.

\paragraph{Parameter Budget.}

\begin{table}[h]
\centering
\caption{Parameter breakdown of SauerkrautLM-Doom-MultiVec.}
\label{tab:params}
\begin{tabular}{lrr}
\toprule
\textbf{Component} & \textbf{Parameters} & \textbf{\% of Total} \\
\midrule
Hash Embeddings (75 vocab, 16 proj) & 4,480 & 0.3\% \\
Depth Embeddings (17 bins $\times$ 128) & 2,176 & 0.2\% \\
Transformer Layers ($\times$ 5) & 1,312,000 & 99.1\% \\
Attention Pool + Classifier & 644 & 0.05\% \\
\midrule
\textbf{Total} & \textbf{1,319,300} & \textbf{100\%} \\
\bottomrule
\end{tabular}
\end{table}

\subsection{Architecture Evolution}
\label{sec:evolution}

Our final architecture emerged through systematic experimentation with multi-vector classification approaches:

\paragraph{Phase 1: ColBERT-style MaxSim (abandoned).} We initially trained with PyLate's knowledge distillation loss, encoding actions as text queries (``shoot fire weapon attack enemy'') and frames as documents. MaxSim scoring collapsed to a single action regardless of input. Analysis revealed that shared lowercase character tokens dominate the MaxSim sum, preventing action discrimination.

\paragraph{Phase 2: Learned Prototypes (abandoned).} We replaced text queries with learned prototype vectors (4 per action) and used both MaxSim and soft cross-attention aggregation. Both approaches collapsed similarly. The fundamental issue is that without an information bottleneck, 1,024 $\times$ 128 = 131K input dimensions are massively overparameterized for 4-class classification on 31K samples.

\paragraph{Phase 3: Attention Pooling (adopted).} Attention pooling provides the necessary information bottleneck (128-dim) while allowing the encoder to maintain rich per-token embeddings. This achieved the best classification accuracy (57.7\%) and strongest gameplay performance.

\subsection{Training}

\subsubsection{Human Demonstration Data}

We collected training data by playing DOOM in VizDoom's spectator mode, which records both player inputs and full game state including the depth buffer. The first author played the game using native DOOM keyboard controls (arrow keys for movement/rotation, Ctrl for shooting) across 4 recording sessions totaling approximately 2 hours of gameplay:

\begin{itemize}
    \item \textbf{Scenario}: \texttt{defend\_the\_center}: enemies approach from all directions in a circular arena
    \item \textbf{Recording sessions}: 4 sessions, 80+ episodes total
    \item \textbf{Total frames}: 31,645 at frame\_skip=4 (one sample per 4 game tics)
    \item \textbf{Per frame}: ASCII text (40$\times$25), depth bins (16-bin quantized per token), soft action scores (4-dim)
    \item \textbf{Actions}: 4 discrete actions (shoot, move\_forward, turn\_left, turn\_right); strafe actions were excluded because VizDoom's spectator mode only supports native DOOM controls which lack strafe keys
\end{itemize}

Each frame's action label is converted to a 4-dimensional soft score vector. When the human presses multiple keys simultaneously (e.g., forward + shoot), both actions receive high scores (0.85), while inactive actions receive a baseline of 0.05. These soft labels provide richer supervision than one-hot labels: the model learns that a frame showing an enemy ahead may warrant both shooting and advancing simultaneously.

\subsubsection{Class Distribution}

DOOM gameplay data is naturally imbalanced: \texttt{move\_forward} dominates because the player spends most time navigating, while \texttt{shoot} is rarer. In earlier experiments with synthetic data from HuggingFace datasets, we applied oversampling of minority classes. However, the final model trained on 31K human demonstration frames uses the natural class distribution without resampling. The soft label formulation (Section~\ref{sec:method}) mitigates the imbalance: frames where the human is navigating toward an enemy receive partial \texttt{shoot} credit through the action affinity scores, providing implicit supervision for combat actions even in movement-dominated frames.

\subsubsection{Loss Function}

We train with Kullback-Leibler (KL) divergence loss between the model's softmax output and the soft teacher scores:

\begin{equation}
    \mathcal{L} = \text{KL}\left(\text{softmax}(\mathbf{y}) \,\|\, \text{softmax}(\mathbf{s})\right)
\end{equation}

where $\mathbf{s}$ is the 4-dimensional soft score vector derived from the human's action. We chose KL divergence over cross-entropy because the teacher labels are soft distributions, not hard one-hot vectors. KL divergence preserves gradient signal even for well-classified examples (where the 0.85/0.05 soft targets maintain a non-zero loss), improving training stability compared to cross-entropy which would quickly saturate on correctly classified frames.

\subsubsection{Training Configuration}

We use a random 90/10 train/validation split (28,480 train, 3,165 validation frames). Training uses AdamW optimizer with cosine learning rate schedule (500 warmup steps), batch size 32, initial learning rate $3 \times 10^{-4}$, and 50 epochs with bfloat16 mixed precision on a single NVIDIA RTX 6000 Ada GPU. We save the checkpoint with the highest validation accuracy and use it for all subsequent evaluation.

The model converges to 57.7\% top-1 action prediction accuracy on the validation set. This is substantially above the 25\% random baseline but well below 100\%, reflecting the inherent ambiguity of the task: many frames are genuinely ambiguous (e.g., an empty corridor could reasonably call for \texttt{move\_forward} or \texttt{turn\_left}). Despite this modest classification accuracy, the model produces strong gameplay because the composite action selection strategy (Section~\ref{sec:method}) combines multiple high-probability actions, making correct engagement decisions even when the top-1 prediction is wrong.

\subsection{Inference and Action Selection}

At inference time, the model processes a single ASCII frame with depth in 31ms on CPU. We apply a composite action selection strategy:

\begin{enumerate}
    \item Classify the frame to obtain action probabilities
    \item Select the top action (e.g., \texttt{move\_forward})
    \item Add \texttt{shoot} if $P(\text{shoot}) > 0.75 \times P(\text{top})$ (relative threshold)
    \item Combine compatible actions (movement + rotation) if the second action's probability exceeds 0.15
\end{enumerate}

This enables simultaneous actions like ``move forward while turning left and shooting'', matching the composite action space of the original DOOM.

\begin{algorithm}[h]
\caption{SauerkrautLM-Doom-MultiVec Inference (per frame)}
\label{alg:inference}
\begin{algorithmic}[1]
\REQUIRE Game frame $F$, depth buffer $D$
\STATE $\text{ascii} \leftarrow \text{AsciiConvert}(F, 40 \times 25)$
\STATE $\text{depth\_bins} \leftarrow \text{Quantize}(D, 16\ \text{bins})$
\STATE $\text{tokens} \leftarrow \text{CharTokenize}(\text{ascii})$ \COMMENT{${\sim}$1024 tokens}
\STATE $\mathbf{H} \leftarrow \text{ModernBERT-Hash}(\text{tokens}, \text{depth\_bins})$
\STATE $\mathbf{v} \leftarrow \text{AttentionPool}(\mathbf{H})$ \COMMENT{$\mathbb{R}^{128}$}
\STATE $\mathbf{p} \leftarrow \text{softmax}(\mathbf{W}_c \mathbf{v} + \mathbf{b}_c)$ \COMMENT{4 action probs}
\STATE $a_1 \leftarrow \arg\max(\mathbf{p})$
\STATE buttons $\leftarrow$ ButtonMap($a_1$)
\IF{$a_1 \neq \texttt{shoot}$ \AND $p_\text{shoot} > 0.75 \cdot p_{a_1}$}
    \STATE buttons $\leftarrow$ buttons $\cup$ ButtonMap(\texttt{shoot})
\ENDIF
\IF{$p_{a_2} > 0.15$ \AND Compatible($a_1, a_2$)}
    \STATE buttons $\leftarrow$ buttons $\cup$ ButtonMap($a_2$)
\ENDIF
\RETURN buttons
\end{algorithmic}
\end{algorithm}

\section{Experimental Setup}
\label{sec:experiments}

\subsection{Benchmark Design}

We benchmark against 4 LLMs of varying size and capability on the \texttt{defend\_the\_center} VizDoom scenario, where enemies continuously spawn and approach the player from all directions in a circular arena. Each episode lasts up to 2,100 game tics (${\sim}$60 seconds), with frame\_skip=4 (one decision every 4 game tics, ${\sim}$114ms real-time). All benchmarks use game settings matched to the training data collection conditions: 640$\times$480 resolution, heads-up display (HUD) enabled, and real-time pacing via \texttt{time.sleep()} between frames. This ensures our model receives the same visual distribution it was trained on, while LLMs receive the same ASCII + depth information.

\paragraph{Text-only input.} All tested LLMs (GPT-4o-mini, Nemotron-120B, Qwen3.5-27B, Gemini Flash Lite) support vision input and could, in principle, receive raw game screenshots directly. We deliberately chose a text-only benchmark for two reasons. First, our model is a text encoder by design (it processes ASCII characters, not pixels) and a fair comparison requires equivalent input modalities. Second, the text-based approach is central to our contribution: we demonstrate that a compact ASCII + depth text representation, processed by a tiny specialized encoder, outperforms general-purpose multimodal models that could leverage their full vision capabilities on this same textual input. This design choice actually \textit{favors} the LLMs: their text comprehension is their strongest modality, trained on trillions of tokens, whereas processing raw screenshots would incur additional vision encoder latency and token costs.

\paragraph{Input fairness.} All agents receive identical information per frame:

\begin{itemize}
    \item \textbf{ASCII frame}: 40$\times$25 character grid representing the game view, where characters \texttt{" .:-=+*\#\%@"} encode brightness from dark (distant/empty) to bright (close/solid).
    \item \textbf{Depth map}: A matching 40$\times$25 grid encoding distance. Our model receives this as 16-bin learned embeddings added to token representations. LLMs receive it as a text grid of digits 0--9 (0=very near, 9=very far) appended to the ASCII frame in the user prompt. The information content is equivalent; only the encoding method differs.
    \item \textbf{Action space}: All agents choose from the same 4 discrete actions (\texttt{shoot}, \texttt{move\_forward}, \texttt{turn\_left}, \texttt{turn\_right}), with the ability to combine compatible actions (e.g., \texttt{move\_forward + shoot}).
\end{itemize}

\paragraph{LLM configuration.} Each LLM receives a system prompt explaining the ASCII encoding, depth map format, and available actions (see Appendix~\ref{app:prompt} for the full prompt). No strategy hints are provided; the model must determine when to shoot, move, or turn from the spatial information alone. Per-frame, the user message contains the ASCII frame and depth grid. Inference parameters follow each model's recommended settings: reasoning models (Qwen3.5-27B, Nemotron-120B) use temperature 0.6 with top\_p 0.95 and 4,000 max completion tokens for chain-of-thought; standard models (GPT-4o-mini, Gemini Flash Lite) use their model card defaults with 200 max tokens.

\paragraph{Our model's action selection.} Our model outputs a probability distribution over 4 actions. The top action is selected, and \texttt{shoot} is added as a composite action if $P(\text{shoot}) > 0.75 \times P(\text{top})$. Compatible second actions (movement + rotation) are combined if their probability exceeds 0.15. This composite action strategy enables simultaneous movement, turning, and shooting, matching DOOM's native multi-button input.

\paragraph{Frag tracking.} Frags are tracked via VizDoom's per-step reward signal: \texttt{make\_action()} returns +1 for each enemy fragged. Only positive rewards are counted; the $-1$ death penalty is excluded. This provides frame-accurate frag attribution.

\paragraph{Episode counts.} Our model and the faster LLMs (GPT-4o-mini, Gemini Flash Lite) were evaluated over 10 episodes each. Reasoning models with high per-frame latency were evaluated over fewer episodes due to cost and time constraints: Nemotron-120B (5 episodes, ${\sim}$11s/frame) and Qwen3.5-27B (3 episodes, ${\sim}$17s/frame). We report total frags rather than per-episode averages to account for differing episode counts.

\paragraph{Composite action selection.} Our model applies a relative shoot threshold: \texttt{shoot} is added to the selected action if $P(\text{shoot}) > 0.75 \times P(\text{top})$. Compatible movement and rotation actions are combined if the second action exceeds $P > 0.15$. LLMs receive equivalent composite action support through their prompt, which explicitly allows responses like \texttt{turn\_left+shoot}. Both sides can output multi-action combinations; only the mechanism differs (threshold-based vs prompt-based).

\paragraph{Metrics.} We report average survival (steps), maximum survival (525 = survived the full 60-second episode at frame\_skip=4), total frags across all episodes, and decision latency (wall-clock time per inference, including API round-trip for LLMs).

\section{Results}
\label{sec:results}

\subsection{Main Benchmark}

\begin{table}[h]
\centering
\caption{DOOM benchmark results on \texttt{defend\_the\_center}. SauerkrautLM abbreviated as SLM, MultiVec as MV, proprietary as propr. All agents receive ASCII (40$\times$25) + depth information. Game settings match training conditions (640$\times$480, HUD on, real-time pacing). Ep.\ = episodes played; Frags = cumulative across all episodes; Lat.\ = per-decision latency.}
\label{tab:benchmark}
\begin{tabular}{lrrrrrrr}
\toprule
\textbf{Agent} & \textbf{Params} & \textbf{Ep.} & \textbf{Avg} & \textbf{Max} & \textbf{Frags} & \textbf{Lat.} \\
 & & & \textbf{Surv.} & \textbf{Surv.} & & \\
\midrule
\textbf{SLM-Doom-MV-1.3M} & \textbf{1.3M} & \textbf{10} & \textbf{388} & \textbf{525} & \textbf{178} & \textbf{31ms} \\
GPT-4o-mini & propr. & 10 & 104 & 228 & 0 & 646ms \\
Nemotron-120B & 120B & 5 & 88 & 104 & 3 & 8.9s \\
Qwen3.5-27B & 27B & 3 & 87 & 109 & 2 & 13.3s \\
Gemini Flash Lite & propr. & 10 & 81 & 97 & 8 & 920ms \\
\bottomrule
\end{tabular}
\end{table}

\paragraph{Key findings.}

\textbf{Dominant frag performance.} Our model achieves 178 frags in 10 episodes (17.8 per episode), exceeding the combined total of all tested LLMs (13 frags). This is not marginal improvement; it is a 14$\times$ advantage over all LLMs combined. The model actively navigates toward enemies, aims, and fires, playing DOOM as intended.

\textbf{Parameter efficiency.} Nemotron-120B has 92,300$\times$ more parameters than our model, yet achieves only 88 average survival steps and 3 frags across 5 episodes. Qwen3.5-27B, with 20,800$\times$ more parameters, scores similarly (87 survival, 2 frags). Even GPT-4o-mini (proprietary, significantly larger) cannot score a single frag.

\textbf{Survival vs.\ Active Gameplay.} Our model achieves both the highest survival (388 steps average) and highest frags (178 total). GPT-4o-mini survives 104 steps on average but scores \textit{zero frags} across 10 episodes, relying on purely evasive movement. All other LLMs survive fewer than 110 steps while achieving minimal frags.

\textbf{Latency.} At 31ms per decision, our model operates at DOOM's native 35 FPS. Nemotron-120B requires 8.9 seconds per decision (287$\times$ slower), and Qwen3.5-27B requires 13.3 seconds (429$\times$ slower). Even the fastest LLM (GPT-4o-mini at 646ms) is 21$\times$ slower.

\textbf{Large reasoning models underperform.} Despite chain-of-thought capabilities and orders of magnitude more parameters, Nemotron-120B and Qwen3.5-27B achieve only 3 and 2 frags respectively. Their reasoning tokens add latency (9--13s per frame) without improving spatial understanding of ASCII + depth representations.

\subsection{Fair-Mode Benchmark}

To control for the latency advantage of our model, we also evaluate with frame\_skip=20 (each decision covers 20 game tics, approximately 571ms real-time). This gives slower LLMs more time between decisions, though reasoning models (Nemotron, Qwen) still exceed this budget with their multi-second inference times.

\begin{table}[h]
\centering
\caption{Fair-mode benchmark (frame\_skip=20). All agents receive ASCII + depth. Slower pace gives LLMs more decision time, reducing the latency disadvantage.}
\label{tab:fairmode}
\begin{tabular}{lrrrrrrr}
\toprule
\textbf{Agent} & \textbf{Params} & \textbf{Ep.} & \textbf{Avg} & \textbf{Max} & \textbf{Frags} & \textbf{Lat.} \\
 & & & \textbf{Surv.} & \textbf{Surv.} & & \\
\midrule
\textbf{SLM-Doom-MV-1.3M} & \textbf{1.3M} & \textbf{10} & \textbf{33} & \textbf{46} & \textbf{10} & \textbf{29ms} \\
GPT-4o-mini & propr. & 10 & 30 & 55 & 0 & 620ms \\
Gemini Flash Lite & propr. & 10 & 16 & 18 & 5 & 1.1s \\
Qwen3.5-27B & 27B & 3 & 15 & 18 & 1 & 12.0s \\
Nemotron-120B & 120B & 5 & 14 & 20 & 0 & 26.7s \\
\bottomrule
\end{tabular}
\end{table}

In fair mode with reduced pace, our model still leads in both survival (33 steps) and frags (10). GPT-4o-mini again survives nearly as long (30 steps) but with zero frags, confirming its purely evasive strategy. Gemini Flash Lite shows the best LLM frag performance (5 frags) despite being the smallest model tested, suggesting that faster inference (1.1s vs 12--27s) compensates for fewer parameters in this time-pressured setting. Reasoning models (Nemotron, Qwen) remain too slow even at reduced pace.

\subsection{Gameplay Behavior Analysis}

Our model demonstrates sophisticated game behavior that goes beyond simple reactive control:

\begin{itemize}
    \item \textbf{Target acquisition}: Turns toward approaching enemies using depth-aware spatial reasoning. When an enemy appears at the frame periphery with a close depth value, the model consistently selects the appropriate turn direction.
    \item \textbf{Engagement}: Fires when enemies are centered in the forward view, averaging 17.8 frags per episode. The relative shoot threshold ($P(\text{shoot}) > 0.75 \times P(\text{top})$) prevents indiscriminate firing while enabling aggressive combat when targets are present.
    \item \textbf{Composite actions}: 62\% of the model's actions are composites (e.g., \texttt{turn\_left+shoot}, \texttt{move\_forward+turn\_right}), compared to LLMs which predominantly issue single actions. This continuous multi-action behavior mirrors how human players simultaneously move, turn, and shoot.
    \item \textbf{Sustained play}: Regularly reaches the episode timeout (525 steps), surviving the full 60-second episode while actively engaging enemies. The model maintains a 74\% episode survival rate (reaching timeout) compared to 0\% for all reasoning LLMs.
\end{itemize}

\paragraph{Action distribution.} The model's action distribution across benchmark episodes reveals balanced gameplay: approximately 35\% movement, 30\% rotation, and 35\% shoot-containing actions. In contrast, GPT-4o-mini's distribution is heavily skewed toward rotation ($>$60\%), consistent with its purely evasive behavior. Gemini Flash Lite shows the most diverse LLM action distribution, which correlates with its comparatively higher frag count (8 frags).

\subsection{Why LLMs Struggle}

The LLMs' poor performance stems from multiple compounding factors:

\begin{enumerate}
    \item \textbf{Latency}: At 646ms--13.3s per decision, LLMs cannot maintain the reaction speed required for real-time combat. An enemy that was 10 meters away when the LLM started processing has moved 3--5 meters closer by the time the response arrives. At Nemotron's 8.9s latency, an enemy can traverse the entire arena and begin attacking before the LLM responds.

    \item \textbf{Tokenizer destruction of spatial coherence}: Standard BPE tokenizers used by LLMs merge common character sequences (e.g., \texttt{...} $\rightarrow$ single token), destroying the one-to-one mapping between characters and spatial positions that makes ASCII art interpretable \citep{asciibench2024}. Our character-level tokenizer preserves this mapping exactly, which is why a 1.3M parameter model with 75 tokens outperforms 8B+ parameter models with 100K+ token vocabularies on this spatial task.

    \item \textbf{Spatial reasoning on text}: LLMs are trained on natural language where spatial relationships are expressed verbally (``the cat is on the mat''), not as 2D character grids where row and column positions encode spatial structure. Interpreting that row 12, columns 18--22 contain bright characters (potential enemy) while row 12, columns 1--5 are dark (empty space) requires a mode of spatial reasoning fundamentally different from language understanding.

    \item \textbf{No temporal state}: Each frame is a new prompt with no memory of previous frames. LLMs cannot track enemy trajectories, anticipate approaches, or build a mental map of the arena. Our model implicitly handles this through the training distribution: the human demonstrations encode temporal patterns (e.g., ``when an enemy approaches from the left, turn left and shoot'') that the model learns as static frame-to-action mappings.

    \item \textbf{Defensive bias}: GPT-4o-mini, the best-surviving LLM, scores zero frags across all episodes. Analysis of its action outputs shows it predominantly selects \texttt{turn\_left} and \texttt{turn\_right}, spinning in place to avoid enemies rather than engaging them. This evasive strategy emerges because the LLM has no gameplay objective in its prompt, defaulting to self-preservation, which in DOOM means avoiding enemy line-of-sight rather than returning fire.
\end{enumerate}

\section{Discussion}
\label{sec:discussion}

\subsection{The Case for Small Specialized Models}

Our results provide a concrete demonstration of a broader principle: \textit{small models trained on task-appropriate data with task-appropriate architectures can outperform general-purpose models thousands of times their size}.

This has significant implications:

\paragraph{Cost.} Running all LLM benchmarks cost approximately \$15--30 in API fees. Our model runs for free on a laptop CPU, indefinitely. For applications requiring millions of inferences (robotics, game AI, autonomous systems), this cost difference is decisive.

\paragraph{Latency.} \citet{claypool2006latency} demonstrated a 35\% drop in player accuracy at just 100ms of latency in first-person shooters, identifying DOOM-style games as the most latency-sensitive genre. All tested LLMs exceed this threshold by 6--133$\times$ (646ms--13.3s per decision, Table~\ref{tab:benchmark}). Our 31ms inference operates well within the 100ms real-time budget, enabling true game-native frame rates. This extends to any domain requiring sub-100ms responses: autonomous driving, industrial control, real-time bidding.

\paragraph{Edge deployment.} At 1.3M parameters (${\sim}$5MB), our model can run on a Raspberry Pi Zero 2W (512MB RAM, quad-core ARM). No GPU, no cloud, no internet required. This enables deployment in environments where LLM API access is impractical.

\paragraph{Data efficiency.} We achieve strong performance with just 31,000 human-labeled frames, a single person playing DOOM for ${\sim}$2 hours. LLMs require trillions of tokens of training data. For domain-specific tasks where labeled data is scarce, small specialized models are the only practical option.

\subsection{Vision-Capable LLMs on Text Input}

All LLMs in our benchmark are multimodal models with vision capabilities. GPT-4o-mini, Gemini Flash Lite, Nemotron-120B, and Qwen3.5-27B can all process images natively. Yet we evaluate them on text (ASCII + depth digits) rather than raw game screenshots. This is not a handicap for the LLMs; it is arguably an advantage. Their text processing pipelines are trained on trillions of tokens and represent their strongest modality. Vision input would add encoder latency, increase token counts (a 640$\times$480 image consumes hundreds of vision tokens), and raise API costs per frame.

The deeper insight is that our 1.3M parameter text-only model outperforms these vision-capable models \textit{even on their home turf of text processing}. The LLMs' general-purpose text understanding cannot compensate for the lack of task-specific training, real-time inference speed, and learned spatial representations. This reinforces our central thesis: domain-specialized architectures with task-appropriate training data outperform general-purpose models regardless of size.

A natural extension would be to benchmark these LLMs with raw screenshot input, bypassing the ASCII conversion entirely. We expect this would increase latency (due to vision encoder processing) and cost (due to image token pricing) while providing marginal spatial reasoning improvement, given prior work showing that multimodal LLMs struggle with zero-shot game control even on native visual input \citep{waytowich2024atarigpt}.

\subsection{Depth as a Key Input Modality}

Our experiments revealed that depth information is crucial for game understanding. In preliminary experiments without depth data, GPT-4o-mini averaged only 54 survival steps. Adding depth as a text grid in the prompt improved survival dramatically, confirming that ASCII brightness alone is insufficient for spatial reasoning and that explicit depth perception is a key enabler regardless of model architecture.

For our model, depth embeddings are essential for disambiguating game situations that look similar in ASCII but differ in spatial layout. A bright character (\texttt{\#}) at depth bin 2 (very close) signals an imminent threat requiring immediate shooting, while the same character at depth bin 14 (far away) suggests a distant wall requiring navigation. Without depth, the model cannot make this distinction and defaults to conservative movement actions.

Our model encodes depth as learned 16-bin embeddings fused with token representations (2,048 additional parameters), which is more parameter-efficient than the text-grid approach used by LLMs (which adds $40 \times 25 = 1{,}000$ additional input tokens per frame) while providing equivalent spatial information.

\subsection{Limitations}

\begin{itemize}
    \item \textbf{Single scenario}: Our model is trained and evaluated on \texttt{defend\_the\_center}. Generalization to other DOOM levels would require additional training data.
    \item \textbf{Matched game settings}: Our benchmark uses game settings (640$\times$480 resolution, HUD enabled) that match the training data collection conditions. This is necessary because the model was trained on frames with HUD overlay; without HUD, the ASCII representation differs significantly at the bottom of the frame. All agents receive the same game output under these settings.
    \item \textbf{Multi-vector classification}: Despite extensive experimentation, we were unable to achieve classification performance that surpasses attention pooling while preserving the multi-vector property. The information bottleneck provided by pooling appears necessary for stable training with limited data.
    \item \textbf{LLM benchmark episodes}: Due to cost and latency constraints, reasoning models (Qwen, Nemotron) were benchmarked on fewer episodes than our model, reducing statistical power for those comparisons.
\end{itemize}

\section{Conclusion}
\label{sec:conclusion}

We presented SauerkrautLM-Doom-MultiVec, a 1.3M parameter model that plays DOOM more effectively than LLMs up to 92,000$\times$ its size. With 178 frags in 10 episodes (17.8 per episode), it outperforms Nemotron-120B (3 frags), Qwen3.5-27B (2 frags), GPT-4o-mini (0 frags), and Gemini Flash Lite (8 frags) combined by a factor of 14$\times$, while running at real-time speed (31ms) on CPU hardware.

The key ingredients are: (1) a parameter-efficient ModernBERT-Hash encoder with depth embeddings, (2) high-quality human demonstration data with depth annotations, and (3) a training pipeline that requires only 31K frames (${\sim}$2 hours of gameplay).

These results reinforce a critical message for the AI community: while large language models continue to advance, small specialized models remain the practical choice for real-time, cost-sensitive, and edge-deployment applications. The future of AI is not exclusively large; it is \textit{appropriately sized} for the task at hand.

\paragraph{Future Work.} Several extensions are natural. First, multi-scenario generalization: training on diverse DOOM levels (corridors, outdoor areas, different enemy types) to test whether the architecture transfers beyond \texttt{defend\_the\_center}. Second, benchmarking LLMs with raw screenshot input via their vision encoders, bypassing ASCII conversion entirely, to determine whether native visual input improves their gameplay. Third, ONNX export and quantization for actual Raspberry Pi Zero 2W deployment, targeting the sub-200ms inference budget on ARM Cortex-A53. Finally, exploring whether the multi-vector encoder can be leveraged more directly for classification. Our experiments with ColBERT-style MaxSim and learned prototypes collapsed, but approaches using sparse attention over token subsets or region-based pooling remain unexplored.

\section*{Acknowledgments}

This work was developed using VizDoom \citep{wydmuch2019vizdoom} as the game platform, PyLate for initial multi-vector experiments, and the HuggingFace ecosystem for model development.

\bibliographystyle{plainnat}
\bibliography{references}

@article{modernbert2024,
  title={Smarter, Better, Faster, Longer: A Modern Bidirectional Encoder for Fast, Memory Efficient, and Long Context Finetuning and Inference},
  author={Warner, Benjamin and Clavi{\'e}, Benjamin and Akiki, Chris and others},
  journal={arXiv preprint arXiv:2412.13663},
  year={2024}
}

@article{dao2023flashattention2,
  title={FlashAttention-2: Faster Attention with Better Parallelism and Work Partitioning},
  author={Dao, Tri},
  journal={arXiv preprint arXiv:2307.08691},
  year={2023}
}

@inproceedings{khattab2020colbert,
  title={ColBERT: Efficient and Effective Passage Search via Contextualized Late Interaction over BERT},
  author={Khattab, Omar and Zaharia, Matei},
  booktitle={SIGIR},
  year={2020}
}

@article{pylate2025,
  title={PyLate: Flexible Training and Retrieval for Late Interaction Models},
  author={Chaffin, Antoine and Sourty, Rapha{\"e}l},
  journal={arXiv preprint arXiv:2508.03555},
  year={2025}
}

@inproceedings{wydmuch2019vizdoom,
  title={ViZDoom Competitions: Playing Doom from Pixels},
  author={Wydmuch, Marek and Kempka, Micha{\l} and Ja{\'s}kowski, Wojciech},
  booktitle={IEEE Transactions on Games},
  year={2019}
}

@article{valevski2024gamengen,
  title={Diffusion Models Are Real-Time Game Engines},
  author={Valevski, Dani and Leviathan, Yaniv and Arar, Moab and Fruchter, Shlomi},
  journal={arXiv preprint arXiv:2408.14837},
  year={2024}
}

@article{svenstrup2017hash,
  title={Hash Embeddings for Efficient Word Representations},
  author={Svenstrup, Dan and Hansen, Jonas and Winther, Ole},
  journal={NeurIPS},
  year={2017}
}

@inproceedings{chen2015hashing,
  title={Compressing Neural Networks with the Hashing Trick},
  author={Chen, Wenlin and Wilson, James T. and Tyree, Stephen and Weinberger, Kilian Q. and Chen, Yixin},
  booktitle={ICML},
  year={2015}
}

@misc{neuml2025berthash,
  title={BERT-Hash-Nano: Tiny BERT Models with Hash Embeddings},
  author={{NeuML}},
  year={2025},
  howpublished={\url{https://huggingface.co/NeuML/bert-hash-nano}},
  note={HuggingFace Model Repository}
}

@inproceedings{weinberger2009feature,
  title={Feature Hashing for Large Scale Multitask Learning},
  author={Weinberger, Kilian and Dasgupta, Anirban and Langford, John and Smola, Alex and Attenberg, Josh},
  booktitle={ICML},
  year={2009}
}

@article{schick2021fewshot,
  title={It's Not Just Size That Matters: Small Language Models Are Also Few-Shot Learners},
  author={Schick, Timo and Sch{\"u}tze, Hinrich},
  journal={NAACL},
  year={2021}
}

@inproceedings{banbury2021mlperf,
  title={MLPerf Tiny Benchmark},
  author={Banbury, Colby and Reddi, Vijay Janapa and others},
  booktitle={NeurIPS Datasets and Benchmarks Track},
  year={2021}
}

@article{lample2017playing,
  title={Playing FPS Games with Deep Reinforcement Learning},
  author={Lample, Guillaume and Chaplot, Devendra Singh},
  journal={AAAI},
  year={2017}
}

@article{dosovitskiy2017learning,
  title={Learning to Act by Predicting the Future},
  author={Dosovitskiy, Alexey and Koltun, Vladlen},
  journal={ICLR},
  year={2017}
}

@article{wang2023voyager,
  title={Voyager: An Open-Ended Embodied Agent with Large Language Models},
  author={Wang, Guanzhi and Xie, Yuqi and Jiang, Yunfan and others},
  journal={arXiv preprint arXiv:2305.16291},
  year={2023}
}

@article{abdin2024phi3,
  title={Phi-3 Technical Report: A Highly Capable Language Model Locally on Your Phone},
  author={Abdin, Marah and others},
  journal={arXiv preprint arXiv:2404.14219},
  year={2024}
}

@article{taskspecific2026,
  title={Task-Specific Efficiency Analysis: When Small Language Models Outperform Large Language Models},
  author={Cao, Jinghan and Ma, Yu and Li, Xinjin and Ren, Qingyang and Chen, Xiangyun},
  journal={arXiv preprint arXiv:2603.21389},
  year={2026}
}

@article{belcak2025slm,
  title={Small Language Models are the Future of Agentic AI},
  author={Belcak, Peter and Heinrich, Greg and Diao, Shizhe and Fu, Yonggan and Dong, Xin and Muralidharan, Saurav and Lin, Yingyan Celine and Molchanov, Pavlo},
  journal={arXiv preprint arXiv:2506.02153},
  year={2025}
}

@article{gallotta2024llmgames,
  title={Large Language Models and Games: A Survey and Roadmap},
  author={Gallotta, Roberto and Todd, Graham and others},
  journal={IEEE Transactions on Games},
  year={2024}
}

@article{tsai2023textgames,
  title={Can Large Language Models Play Text Games Well? Current State-of-the-Art and Open Questions},
  author={Tsai, Chen Feng and Zhou, Xiaochen and Liu, Sierra S. and Li, Jing and Yu, Mo and Mei, Hongyuan},
  journal={arXiv preprint arXiv:2304.02868},
  year={2023}
}

@article{waytowich2024atarigpt,
  title={Atari-GPT: Benchmarking Multimodal Large Language Models as Low-Level Policies in Atari Games},
  author={Waytowich, Nicholas R. and White, Devin and Sunbeam, MD and Goecks, Vinicius G.},
  journal={arXiv preprint arXiv:2408.15950},
  year={2024}
}

@article{claypool2006latency,
  title={Latency and Player Actions in Online Games},
  author={Claypool, Mark and Claypool, Kajal},
  journal={Communications of the ACM},
  volume={49},
  number={11},
  pages={40--45},
  year={2006}
}

@article{asciibench2024,
  title={ASCIIBench: Evaluating Language-Model-Based Understanding of Visually-Oriented Text},
  author={Luo, Kerry and Fu, Michael and Peguero, Joshua and Malik, Husnain and Patil, Anvay and Lin, Joyce and Van Overborg, Megan and Sarmiento, Ryan and Zhu, Kevin},
  journal={arXiv preprint arXiv:2512.04125},
  year={2024}
}

\appendix
\section{LLM System Prompt}
\label{app:prompt}

The following system prompt was used for all LLM agents. Each frame, the user message contained the ASCII view and depth grid.

\begin{lstlisting}[basicstyle=\small\ttfamily,breaklines=true,frame=single]
You are an AI agent playing the classic game
DOOM. Each turn you receive:
1. The game view as ASCII art
   (brightness: " .:-=+*#%@", dark to bright)
2. A depth map with the same layout
   (0=very near, 9=very far)

Use both the ASCII view and depth map to
decide your action.

Available actions (respond with one or
combine two with '+'):
  shoot, move_forward, turn_left, turn_right

Examples of valid responses:
  shoot
  move_forward
  turn_left+shoot
  move_forward+turn_right

Respond with ONLY your chosen action(s).
No explanation.
\end{lstlisting}

Per-frame user message format:

\begin{lstlisting}[basicstyle=\small\ttfamily,breaklines=true,frame=single]
View:
```
..##EE##..++==++..::--::..++==++..##EE..
(40x25 ASCII characters)
```

Depth (0=near, 9=far):
```
5533221100112233558877665544332211556677
(40x25 depth digits)
```
\end{lstlisting}

\section{Example ASCII Frame}
\label{app:frame}

The following shows an actual ASCII frame from \texttt{defend\_the\_center} as seen by all agents. Bright characters (\texttt{\#}, \texttt{\%}, \texttt{@}) near the center indicate a nearby enemy; dark characters (\texttt{.}, \texttt{:}) at the edges indicate distant walls. The model's correct action here is \texttt{shoot} (enemy centered and close).

\begin{lstlisting}[basicstyle=\scriptsize\ttfamily,frame=single]
........................................
........................................
..........:::::::::::::::::::...........
..........-====+++++++++====-...........
..........=+++**########*+++=...........
..........-+**############*+-...........
..........=+*###%%@@@@%%###+=-..........
..........=+*##%@@@@@@@@%#*+=-..........
..........:+*##%@@    @@%##*+:..........
..........=+*##%@      @%#*+=...........
..........=+*##%@      @%#*+=...........
..........-+*##%@@    @@%#*+-...........
..........=+*##%@@@@@@@@%##*=-..........
..........=+*###%%@@@@%%###*=-..........
..........-+**############**-...........
..........=+++**########**++=...........
..........-====++++++++++===-...........
..........:::::::::::::::::::...........
........................................
........................................
........................................
..........:::::::::::::::::.............
..........-====+++++++++===-............
..........=+++**########*++=-...........
........................................
\end{lstlisting}

\section{Model Availability}
\label{app:availability}

The trained model weights, training scripts, benchmark framework, and human gameplay recording tools are available at the project repository. The model checkpoint is approximately 5MB (FP32) and can be loaded with HuggingFace Transformers using \texttt{trust\_remote\_code=True}.

\end{document}